\pdfoutput=1
\documentclass[review]{elsarticle}
\usepackage{subfigure}
\usepackage{longtable}
\usepackage{bm}
\usepackage{booktabs}
\usepackage{multirow}
\usepackage{geometry}
\usepackage{amsfonts,amssymb}
\usepackage[linesnumbered,ruled,vlined]{algorithm2e}
\usepackage{algorithmic}

\usepackage{lineno,hyperref}
\modulolinenumbers[5]

\journal{arXiv}

\begin{document}

\begin{frontmatter}

\title{Selecting Related Knowledge via Efficient Channel Attention for Online Continual Learning}


\author[mymainaddress]{Ya-nan Han}
%\ead{luning_zhang@foxmail.com}

\author[mymainaddress]{Jian-wei Liu\corref{mycorrespondingauthor}}
\cortext[mycorrespondingauthor]{Corresponding author}
%\ead{liujw@cup.edu.cn}

\address[mymainaddress]{Department of Automation, College of Information Science and Engineering,
China University of Petroleum , Beijing, Beijing, China}

\begin{abstract}%摘要
Continual learning aims to learn a sequence of tasks by leveraging the knowledge acquired in the past in an online-learning manner while being able to perform well on all previous tasks, this ability is crucial to the artificial intelligence (AI) system, hence continual learning is more suitable for most real-word and complex applicative scenarios compared to the traditional learning pattern. However, the current models usually learn a generic representation base on the class label on each task and an effective strategy is selected to avoid catastrophic forgetting. We postulate that selecting the related and useful parts only from the knowledge obtained to perform each task is more effective than utilizing the whole knowledge. Based on this fact, in this paper we propose a new framework, named Selecting Related Knowledge for Online Continual Learning (SRKOCL), which incorporates an additional efficient channel attention mechanism to pick the particular related knowledge for every task. Our model also combines experience replay and knowledge distillation to circumvent the catastrophic forgetting. Finally, extensive experiments are conducted on different benchmarks and the competitive experimental results demonstrate that our proposed SRKOCL is a promised approach against the state-of-the-art.
\end{abstract}

\begin{keyword}%关键词
Continual learning, related knowledge, Catastrophic forgetting, Knowledge distillation
\end{keyword}

\end{frontmatter}

\section{Introduction}%正文

While deep learning techniques have realized tremendous successes in a wide range of applications over the last decade, most approaches mainly focus on single, isolated problems where the model need to be retrained or redesigned from scratch when they face a new task. As a result, when working in non-stationary environment where stream of instances involves different learning task, then a traditional deep neural network will face catastrophic forgetting of previously learned knowledge \cite{1french2006catastrophic}. The aim of continual learning is to learn consecutive tasks by leveraging the knowledge obtained in the past while should not forget how to perform previously learned tasks. 

In the current literature, most of exist continual learning methods are limited to this situation where the whole task data are observed at each step and the agent would learn the current task by many epochs. However, such learning methods are poorly scalable for most real-word scenarios, where data arrives in a stream style and the agent need to rapidly learn new skills from the current task. In this work, we would develop our continual learning approach in online scenario where every task would be learned in an online manner with training data coming sequentially. Optimizing DNN with such a regime is more practical and appealing and requires many training episode and various techniques in this process while struggling when data comes in a single epoch setting \cite{2sahoo2018online,3pham2019bilevel}.

In general, in order to effectively learn in online continual learning setup, the model should be able to avoid catastrophic forgetting of previously learned information and can perform well on the all tasks at the end of learning. While the initial successes have been achieved recent years, such as regularization-based methods \cite{4kirkpatrick2017overcoming,5zenke2017continual,6leibe2016computer}, architecture-based methods \cite{7yoon2018lifelong,8rusu2016progressive}, etc, there are still a lot of open problems that need to be addressed. Regularization-based continual learning methods compute the important weight parameters for every task and control their change when facing the new task. However, these kinds of method have limited ability to overcome catastrophic forgetting, especially in online setting. In contrast, architecture-based methods avoid catastrophic forgetting by increasing the capacity of neural network. Obviously, these methods are not well suitable for a large number of tasks since they need to a large number of memories for every task to alleviate the forgetting \cite{9parisi2019continual}. 

In addition, a favor latent representation can effectively improve the performance of the model especially in online setting. Unfortunately, most of these methods learn a generic representation by using the whole knowledge. Instead of this, we consider that learning to pick only the relevant beneficial knowledge from the prior knowledge to perform current task instead of using all the previously learned knowledge is more reasonable. This is inspired by human learning. For example, a student from computer science major with a mathematical background should be more preferable to learn the prerequisite courses such as data structure, database system, digital signal processing, etc. This basic mathematical knowledge is can help to learn professional knowledge where one selects the useful information from everyone’s mathematical background according to the context of each course. 

Inspired from above discussion, we propose a framework, called Selecting Related Knowledge for Online Continual Learning (SRKOCL). The SRKOCL can learn multiple tasks sequentially with neural network while not forgetting the knowledge obtained from the old tasks via experience replay strategy and pool knowledge distillation strategy. 

More specifically, jointly training via episodic memory is employed to avoid catastrophic forgetting. However, during the jointly training with previous data stored from previous tasks, the imbalance between the previous and current data is another crucial challenge, which can cause a bias towards current task. To overcome this challenge, our SRKOCL uses pooled outputs distillation \cite{10douillard2020podnet} technology to avoid this, which can match global statistics at various feature levels between the previous and current models to further enforce a suitable trade-off between the old and new knowledge. In addition, we propose to apply an efficient channel attention mechanism to automatically select the most relevant representation from the generic feature bank to improve the classification accuracy. Finally, experimental comparisons are introduced to assess the proposed SRKOCL approach. 

To sum up, our main contributions in this work are listed as follows:

1) We proposed a novel SRKOCL framework in online continual learning scenario where this framework can learn a sequence of task while not forgetting the old knowledge learned from old tasks.

2) Consider that only part of the relevant knowledge is useful for the current task, so we propose to apply an efficient channel attention mechanism to learn to automatically pick the useful knowledge to perform current task.

3) To this end, we perform comprehensive experiments on several commonly used datasets in continual learning literature to validate the effectiveness of our proposed SRKOCL algorithm against a series of state-of-the-art algorithms from continual learning, and the comparative experiment results illustrate that our SRKOCL algorithm is a promising approach for continual learning.

\section{Related Work}%正文

Continual learning, also called lifelong learning, is a long standing research direction in the field of machine learning, which considers learning multiple task sequentially where the agent has to learn the new knowledge from the current task while avoiding to forget previous knowledge learned from past tasks \cite{9parisi2019continual}. Roughly speaking, the existing works for continual learning can be broadly divided into three categories: regularization-based, architecture-based and memory-based.

In order to avoid catastrophic forgetting, the regularization-based approaches such as Elastic Weight Consolidation(EWC)\cite{4kirkpatrick2017overcoming}, Synaptic Intelligence(SI) \cite{5zenke2017continual}, etc, typically attempt to employ a regularization term to the objective function and then preserve as much as possible the weights that are important to perform previous tasks. The differences among these works lie in the weight importance measurement. In other relevant work \cite{6leibe2016computer,11rebuffi2017icarl,12he2020incremental}, regularization is applied to penalize the feature space on task learned in the past to preserve the knowledge in the original model. While these works can obtain a good solution for whole tasks, such methods learn a generic representation only using class label for each task and ignore the fact that there is part of useful knowledge is to the current task. 

These types of architecture-based methods attempt to change architectural properties in response to new knowledge, and then the old information on already learned task is maintained by this way. For example, Rusu et al.\cite{8rusu2016progressive} develop progressive neural which dynamically expand the architecture by assigning new sub-module with fix capacity to be learned on new tasks and freezing the modules trained on past tasks. Unfortunately, this strategy that grows the network to avoid forgetting in continual learning setting would be not scalable for a large scale tasks due to their computational cost and a fixed capacity that the assumption is unrealistic \cite{9parisi2019continual}.

These memory-based approaches alleviate catastrophic forgetting by storing previous experience explicitly while the old instance stored in memory would be used for rehearsal. For instance, early researchers \cite{13robins1993catastrophic} try to alleviate forgetting of previously knowledge using experience memory which is replayed regularly, and such approaches are still utilized today \cite{11rebuffi2017icarl}. In other works, episodic memory is applied as a constraint to mitigate catastrophic forgetting while allowing beneficial transfer of knowledge to past tasks, such as GEM \cite{14lopez2017gradient}, Rwalk \cite{15chaudhry2018riemannian}. 

\section{Problem Formulation}

Following the recent continual learning literature \cite{14lopez2017gradient,3pham2019bilevel}, for the online continual learning setting, a model needs to learn new tasks sequentially from an online data stream while each instance is observed only once. Classically, a deep model $f:R{{\kern 1pt} ^{\;d}}\, \to \;\,R{{\kern 1pt} ^{\;c}}$, parameterized by a vector $\theta \;$,  is utilized. When encountering the input instance $x\;_i^t$  from the task $t$,  we should make the formula ${f_\theta }\left( {x\,_i^t} \right) \approx y\,_i^t$ holds for each instance. Furthermore, we adopt the multi-head leaning setup, that means, the classifier has to access to task identity. The model aims to train classifier to learn sequential tasks from data stream while not forgetting the knowledge obtained from previous tasks.

\section{The Proposed SRKOCL Framework}

The final goal of continual learning is to mimic human learning. Catastrophic forgetting is the long-standing challenge in continual learning, so the starting point for this setting to achieve this goal is to overcome catastrophic forgetting, that means, the model should keep the previous knowledge learned from old tasks when learning new knowledge from current task. In addition, the model should learn to pick the related knowledge from the previously obtained knowledge to complete the learning of current task. In this section, we describe our proposed SRKOCL approach in detail that aim to address these two problems. Fig. 1 shows our proposed SRKOCL framework, and this model is mainly composed of three parts: experience replay, pooled outputs distillation and efficient channel attention mechanism. More concretely, experience replay via episodic memory is used to overcome catastrophic forgetting, and pooled outputs distillation is applied to prevent the data imbalance between the current and the previous task during jointly training. Then, we apply an efficient channel attention mechanism to automatically pick the useful and related knowledge to improve the performance of the model. The details of each phase are illustrated in the following paragraphs.

\textbf{Experience Replay (ER).}Compare to the other types of methods, such as the regularization-based approaches, etc, the memory-based approaches show great advantages. Recently, ER has been widely applied to avoid catastrophic forgetting in continual learning task. For ER strategy we utilize, we conduct two simple modifications. Firstly, a subset of the instances from prior tasks is stored by a memory buffer ${\cal M}$ with the size of $M$, where we select ring buffer as the writing strategy which can ensure the balance between all classes of each task. Then, when the incoming minibatch ${B_n}$ is observed, we concatenate ${B_n}$ with another minibatch ${B_{\cal M}}$ of instances obtained from the memory buffer ${\cal M}$. After that, an SGD step with combined batch is performed to update our model. Finally, we select cross entropy as our loss function and the training objective can be formulated as following:

\begin{equation}
{L_{pre}}\left( {\theta ,B_n^t \cup B_m^{er}} \right) =  - \mathbb{E}_{\left( {{x_{\kern 1pt} }_i^k,\;{y_{\kern 1pt} }_i^k} \right){\kern 1pt} \, \cup \,{{\cal M}_{\kern 1pt} }^{1\;:\;k}}\sum\limits_{c = 1}^C {\;{{\bf{1}}_{\left[ {c\; = \;{y_{\kern 1pt} }_i^k} \right]}}} \log \left( {\sigma \left( {{f_\theta }\left( {{x_{\kern 1pt} }_i^k} \right)} \right)} \right.{\rm{ }}
\end{equation}

\begin{figure}[!htbp]
	\label{fig1}
	\centering
	\subfigure{
		\includegraphics[width = .9\textwidth]{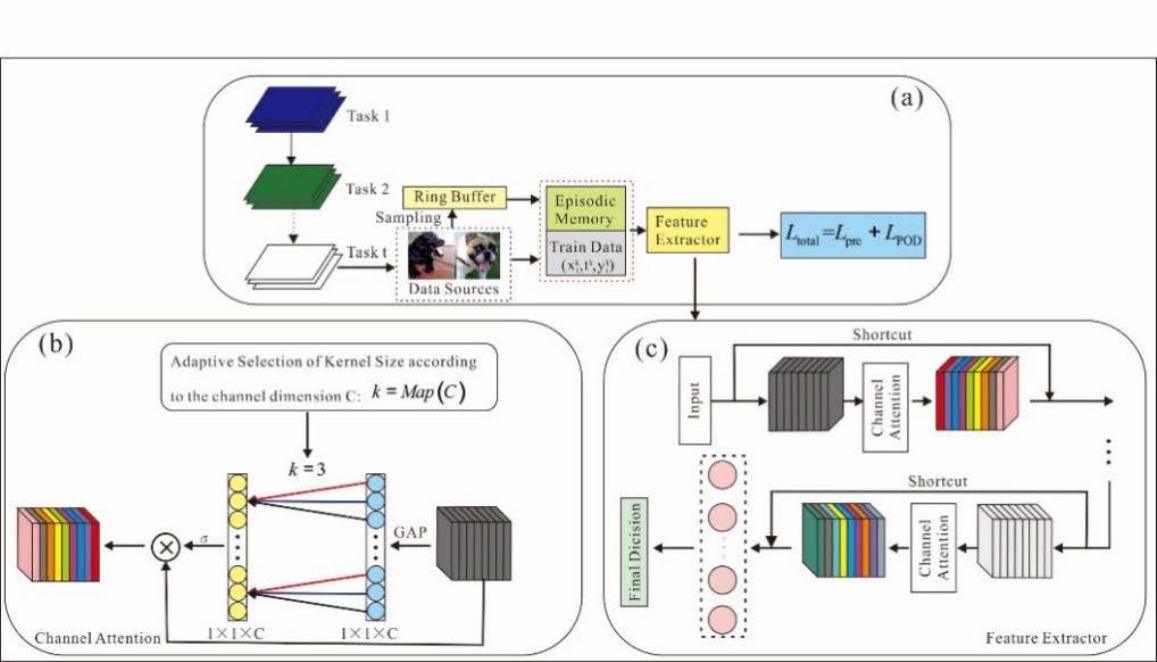}
	}
	\caption{The illustration of the novel SRKOCL network. Above (a): shows SRKOCL at training time. When the task t arrives, the latent representation is learned via a feature extractor (such as ResNet) and here  denotes loss function. To prevent forgetting, the neural network is retained via experience replay with an episodic memory which is written by ring buffer. Right bottom(c): is the feature extractor which is used to learn the latent representation. Left bottom (b): denotes the efficient channel attention module added after each convolutional layer which can select the useful and relevant knowledge instead of taking the whole knowledge.}
\end{figure}

\textbf{Pooled Outputs Distillation (POD).}Consider that in the training process the current and previous task data are used to joint training simultaneously. However, compared to the current task data, previous task data are limited in this step, which would lead to an imbalance between the current and the previous task data. Such imbalance will make this model more favorable to learn the current task, leading to performance degradation. Therefore, in this work we apply POD technology \cite{10douillard2020podnet} to remove the imbalance and further make a trade-off between the new and old knowledge. 

Specifically, POD technology consists in minimizing the difference of global statistics at various feature levels between the new and old model. Let $z$ denote an embedding tensor of size $H \times W \times C$ and then, We extract a POD embedding $\Phi $, which consists of two complementary intermediate statics: the  $W \times C$ height-pooled slices and the $H \times C$ width-pooled slices of the embedding tensor $z$:

\begin{equation}
\Phi \left( z \right) = \left[ {\frac{1}{W}\sum\limits_{w = 1}^W {z\left[ {:,w,:} \right]\left\| {\frac{1}{H}\sum\limits_{h = 1}^H {z\left[ {h,:,:} \right]} } \right.} } \right] \in {R^{\left( {H + W} \right) \times C}}
\end{equation}

where $\left[ { \cdot \;\left\| {\; \cdot } \right.} \right]$ denotes concatenation over the channel axis, and then the differences for both the previous and current model are computed at several layers( in fact, each stage of a ResNet \cite{16he2016deep}). Then, we selected L2 distance between the two sets of embedding as the POD loss function:

\begin{equation}
{L_{pod}}\left( \theta  \right) = \frac{1}{L}\sum\limits_{l = 1}^L {\;{{\left\| {\Phi \left( {f_l^t} \right) - \Phi \left( {f_l^{t - 1}} \right)} \right\|}^2}} 
\end{equation}

\textbf{Efficient Channel Attention Mechanism.}In general, the learned previous knowledge is shared among all tasks. So, the model should enjoy the ability that can automatically pick the useful information from the generic feature bank \cite{17sokar2021self}. To address this problem, we proposed to apply efficient channel attention \cite{18wang2020eca} to make the network to recalibrate the useful and relevant knowledge based on the input instances. In particular, an attention block is employed after each convolutional layer and the role of this block is to recalibrate adaptively the convolutional channels. After that, this model can learn to boost the useful informative features according to the input. Let $z = \left\{ {{z_1},{z_2}, \cdots ,{z_c}} \right\} \in {R^{H \times W \times C}}$ denotes the convolutional outputs and $H$, $W$ and $C$ denote height, width and depth of the feature maps. The vector $s$ of size $C$ denotes the output of every attention block, which can recalibrate the feature map ${z_i} \in {R^{H \times W}}$  for each channel.

\begin{equation}
{z_i^{'}}= {z_i} \circ {s_i}
\end{equation}

where ${s_i}$ denotes the scalar value for channel $i$ and $ \circ $ denotes a channel-wise multiplication.

Note that channel and its weight need a direct correspondence while avoiding dimensionality reduction is more important than consideration of nonlinear channel dependencies \cite{18wang2020eca}. In order to capture local cross-channel interaction and ensure both efficiency and effectiveness, the weight of ${z_i}$  is computed by only considering interaction between ${z_i}$ and its $k$  neighbors:

\begin{equation}
{s_i} = \sigma \left( {\sum\limits_{j = 1}^k {w_i^jz_i^j} } \right),\;z_i^j \in \Omega _i^k
\end{equation}

Where $\Omega \;_i^k$ denotes the set of $k$ adjacent channels of ${z_i}$. Making the entire channels to share the same parameters is more efficient, as is shown in Fig.1, that means:

\begin{equation}
{s_i} = \sigma \left( {\sum\limits_{j = 1}^k {{w^j}z_i^j} } \right),\;z_i^j \in \Omega _i^k
\end{equation}

This process can be achieved by a fast 1D convolution with kernel size $k$,

\begin{equation}
s = \sigma \left( {{\rm{C1D}}\left( z \right)} \right)
\end{equation}

where C1D denotes 1D convolution. It is noted that the kernel size of 1D convolution need to be determined. Inspired by group convolutions \cite{19ioannou2017deep}, high-dimensional(low-dimensional) channels should involve long-range(short-range) convolutions when the number of groups is given. So, it is reasonable that the kernel size $k$ has proportional to channel dimension $C$, that means, there might be a mapping $C = \phi \left( k \right)$.  In addition, consider that we usually set the channel dimension $C$ to power of 2. Therefore, based on this fact, we extend the simple linear function to a non-linear situation,

\begin{equation}
C = \phi \left( k \right) = {2^{\left( {\lambda  \cdot k + b} \right)}}
\end{equation}

After that, the kernel size $k$ can be automatically determined by the following Eq.(9) when the channel dimension $C$ is known, 

\begin{equation}
k = {\left| {\frac{{{{\log }_2}\left( C \right)}}{\lambda } + \frac{b}{\lambda }} \right|_{odd}}
\end{equation}

Noted that in this work the values of $\lambda $ and $b$ are set to 2 and 1 respectively. Clearly, by computing the Eq.(9), we can achieve that the longer range interaction is involved when the channel dimension is higher while low-dimensional channels have shorter range interaction.

\section{Formulating Whole Model}

In this section, we would illustrate our whole model. Our model incorporates a classical convolutional network (in fact, a reduced ResNet) as a feature extractor, and a classifier is used to make predictions. Then, we introduce three parts to our SRKOCL model: (1) the experience replay is applied to prevent the catastrophic forgetting; (2) the knowledge distillation ${L_{pod}}$ is to make an trade-off between the current knowledge and the previous knowledge to remove the imbalance between the current task and the previous task data; (3) we devise an efficient channel attention mechanism to adaptively select the relevant knowledge from the old knowledge instead of using the entire knowledge. Finally, the ultimate loss is formulated as follows:

\begin{equation}
{L_{total}} = {L_{pre}} + {L_{pod}}
\end{equation}

The objective function can be divided into two parts: the cross-entropy loss and the pooled output distillation. Our proposed SRKOCL algorithm is illustrated in Algorithm 1 in details.

\begin{algorithm}
	\renewcommand{\algorithmicrequire}{\textbf{Initialize:}}
	\renewcommand{\algorithmicensure}{\textbf{Require:}}
	\caption{SRKOCL algorithm}
	\label{alg1}
	\begin{algorithmic}[1]
		\REQUIRE The model parameters $\theta $, episodic memory ${{\cal M}_{\kern 1pt} } \leftarrow \left\{ {} \right\}$ and learning  Rate: $\eta $
		\ENSURE memory update strategy for  ${{\cal M}_{\kern 1pt} }$ \\
           \For{$k = 0$ \KwTo $T$}{
          \emph {The train dataset ${D_k}$  comes sequentially}\\
           \For{$n = 0$ \KwTo ${N_{batches}}$}{
    
       \emph {Observe a mini-batch of training instances ${B_n}$ from the current task $k$}\\
        \emph{Compute the total loss ${L_{\;total}}$ using $\left( {x,y} \right) \in {B_n} \cup {B_{\cal M}}_{\kern 1pt} $  by Eq.(10)}\\
       \emph{Update the model parameters $\theta  \leftarrow \theta  - \eta {\kern 1pt} \Delta {\kern 1pt} {\kern 1pt} {L_{\;total}}$\\}
          }
%    \EndFor
    \emph { ${{\cal M}_{\kern 1pt} } \leftarrow Update\_Memory\left( {mem\_size,\;k,\;{{\cal M}_{\kern 1pt} },\;{B_n}} \right)$ } \\ 
    
	\Return $\theta ,{\kern 1pt} \;{{\cal M}_{\kern 1pt} }$ }
	\end{algorithmic}
\end{algorithm}

%\IncMargin{1em}
%\begin{algorithm} \SetKwData{Left}{left}\SetKwData{This}{this}\SetKwData{Up}{up} \SetKwFunction{Union}{Union}\SetKwFunction{FindCompress}{FindCompress} \SetKwInOut{Input}{input}\SetKwInOut{Output}{output}
%	
%	\renewcommand{\algorithmicrequire}{\textbf{Initialize:}}
%	\renewcommand{\algorithmicensure}{\textbf{Require:}}
%	% \BlankLine 
%	 \REQUIRE The model parameters $\theta $, episodic memory ${{\cal M}_{\kern 1pt} } \leftarrow \left\{ {} \right\}$ and learning  Rate: $\eta $
%	\ENSURE memory update strategy for  ${{\cal M}_{\kern 1pt} }$
%	 \emph{special treatment of the first line}\; 
%	 \For{$i\leftarrow 2$ \KwTo $l$}{ 
%	 	\emph{special treatment of the first element of line $i$}\; 
%	 	\For{$j\leftarrow 2$ \KwTo $w$}{\label{forins} \Left$\leftarrow$\FindCompress{$Im[i,j-1]$}\; 
%	 	\Up$\leftarrow$ \FindCompress{$Im[i-1,]$}\; 
%	 	\This$\leftarrow$ \FindCompress{$Im[i,j]$}\; 
%	 	\If(\tcp*[h]{O(\Left,\This)==1})
%	 		{\Left compatible with \This}{\label{lt} 
%	 			\lIf{\Left $<$ \This}{\Union{\Left,\This}}
%	 			 \lElse{\Union{\This,\Left}} } 
% 	    \If(\tcp*[f]{O(\Up,\This)==1}){\Up compatible with
% 			 	 \This}{\label{ut} 
% 			 	 \lIf{\Up $<$ \This}{\Union{\Up,\This}} 
% 			 	 \tcp{\This is put under \Up to keep tree as flat as possible}\label{cmt} \lElse{\Union{\This,\Up}}\tcp*[h]{\This linked to \Up}\label{lelse}
% 		 	  }
% 	 	   }
% 		 \lForEach{element $e$ of the line $i$}{\FindCompress{p}} 
% 	 } 
% 	 	  \caption{disjoint decomposition}
% 	 	  \label{algo_disjdecomp} 
% 	 \end{algorithm}
% \DecMargin{1em} 

\section{Experiments}

In this section, firstly, we review the benchmark datasets used in our evaluation as well as baselines we compared against. Next, we describe the implementation details. Finally, we perform a set of experiments to validate the efficacy of our SRKOCL, and then, we report and analysis the experimental results obtained on different baseline datasets.

\subsection{Datasets and Baselines}

\textbf{Datasets.} The following benchmarks are used to evaluate our proposed SRKOCL algorithm. \textbf{Split CIFAR100} is a variant of the CIFAR100 dataset \cite{20krizhevsky2009learning}, where every task consists of 5 various classes without any replacement from the total of 100 classes. We split CIFAR100 dataset into 20 disjoint subsets, and each of which can be regarded as a separated task, that means in this dataset there are 20 tasks in total; \textbf{Split CIFAR10} divide the CIFAR-10 dataset \cite{20krizhevsky2009learning} into 5 various tasks and 2 classes included in every task, similarly as Split CIFAR100; \textbf{Split SVHN \cite{21rajasegaran2020itaml}} includes 600,000 digit images from Goodle Street View and we also divide this dataset into 5 disjoint tasks and 2 classes contained in each task.

\textbf{Baselines.} We compare our SRKOCL algorithm to the following state-of-the-art approaches: \textbf{LwF} (Learning without Forgetting) \cite{6leibe2016computer}, \textbf{EWC}(Elastic Weight Consolidation) \cite{4kirkpatrick2017overcoming}, \textbf{ICARL} (Incremental Classifier and Representation Learning) \cite{11rebuffi2017icarl}, \textbf{GEM} (Gradient Episodic Memory) \cite{14lopez2017gradient}, \textbf{ER} (Experience Replay) \cite{22chaudhry2019continual}, \textbf{FTML} (Following the Meta Learner) \cite{23finn2019online}.

\subsection{Implementation Datails}

We follow the implementation details proposed in  \cite{14lopez2017gradient,3pham2019bilevel} in all of our experiments. Specifically, we use a reduced ResNet18 in our all experiments. In addition, we apply stochastic gradient decent (SGD) to optimize our model with the bath-size 10 for all approaches in online learning fashion. We use the experience replay strategy to avoid catastrophic forgetting and the size of episodic memory is set to 65 and 25 for every task on CIFAR and SVHN datasets respectively. For each model, we conduct experiments five times and the average numerical results are reported in this paper.

\subsection{Results and Analysis}

In this subsection, we perform a variety of experiments to validate the effectiveness of SRKOCL in continual learning scenario. In the first set of experiments, we compare the performance of our SRKOCL algorithm to the state-of-the-art approaches on different benchmark datasets in continual learning literatures via three evaluation metrics: average accuracy(ACC), forgetting measure (FM) \cite{14lopez2017gradient}, learning accuracy(LA) \cite{24riemer2018learning}. The average accuracy denotes the classify performance on all of tasks when the learner finishes training on the last task $T$. Forgetting is used to evaluate the model’s ability to maintain previous knowledge when the new knowledge is obtained, that means, the smaller this metric, the less forgetting. Lastly, learning accuracy is applied to measure the model’s ability to learn quickly.

Next, we offer more insight for our SRKOCL and conduct an ablation study to evaluate the contribution of each component in this algorithm on different datasets. The final numerical results are reported in Table 1. Here, \textbf{SRKOCL-Base}: the model is trained without efficient channel attention mechanism and feature knowledge distillation, \textbf{SRKOCL-POD}: the model is trained using feature knowledge distillation alone, \textbf{SRKOCL}: the model is trained using feature knowledge distillation and efficient channel attention mechanism.

\begin{table}[!htbp]
	\centering
	\caption{Experimental results of different algorithms on different datasets}
	\begin{tabular}{cccc}
		\hline
		\multirow{2}{*}{Method} & \multicolumn{3}{c}{Split CIFAR10(5T)}\\ \cline{2-4}
	&ACC($ \uparrow $)	 &FM($ \downarrow $) &LA($ \uparrow $)	
\\ \hline
LWF	&0.7574$ \pm $0.0216	 &0.1627$ \pm $0.0308	&0.8875$ \pm $0.0186 \\
EWC	&0.7398$ \pm $0.0496	&0.1873$ \pm $0.0510	 &0.8794$ \pm $0.0338 \\
ICARL	 &0.8057$ \pm $0.0249
	&0.1181$ \pm $0.0313	 &0.9002$ \pm $0.0036 \\
GEM	&0.8491$ \pm $0.0121	&0.0585$ \pm $0.0208	&0.8940$ \pm $0.0093 \\
ER	&0.8465$ \pm $0.0174	 &0.0600$ \pm $0.0151	&0.8945$ \pm $0.0121 \\
FTML	&0.8400$ \pm $0.0291	 &0.0779$ \pm $0.0446	 &0.8998$ \pm $0.0155 \\ \hline
\textbf{SRKOCL-Base}	&0.8537$ \pm $0.0109	 &0.0619$ \pm $0.0163	&0.9025$ \pm $0.0095 \\
\textbf{SRKOCL-POD}	&0.8726$ \pm $0.0195	 &0.0365$ \pm $0.0238	&0.8919$ \pm $0.0145 \\
\textbf{SRKOCL}	&0.8876$ \pm $0.0078	 &0.0320$ \pm $0.0053	&0.9035$ \pm $0.0124 \\ \hline
	\end{tabular}
\end{table}

\begin{table}[!htbp]
	\centering
	\begin{tabular}{cccc}
		\hline
		\multirow{2}{*}{Method} & \multicolumn{3}{c}{Split SVHN(5T)}\\ \cline{2-4}
	&ACC($ \uparrow $)	 &FM($ \downarrow $) &LA($ \uparrow $)	
\\ \hline
LWF	&0.8327$ \pm $0.0196	 &0.1337$ \pm $0.0365	&0.9397$ \pm $0.0111 \\
EWC	&0.8265$ \pm $0.0179	 &0.1466$ \pm $0.0410	&0.9439$ \pm $0.0160 \\
ICARL	&0.8689$ \pm $0.0204
	&0.1062$ \pm $0.0354	&0.9539$ \pm $0.0104 \\
GEM	&0.9000$ \pm $0.0159	 &0.0714$ \pm $0.0308	&0.9572$ \pm $0.0102 \\
ER	&0.9157$ \pm $0.0137	&0.0608$ \pm $0.0182	&0.9643$ \pm $0.0041  \\
FTML	&0.9330$ \pm $0.0063	&0.0422$ \pm $0.0103	&0.9668$ \pm $0.0021 \\ \hline
\textbf{SRKOCL-Base}	&0.9215$ \pm $0.0167 	&0.0511$ \pm $0.0256	  &0.9616$ \pm $0.0051 \\
\textbf{SRKOCL-POD}	&0.9396$ \pm $0.0047	&0.0216$ \pm $0.0068	&0.9529$ \pm $0.0081 \\
\textbf{SRKOCL}	&0.9410$ \pm $0.0048	&0.0169$ \pm $0.0116	&0.9527$ \pm $0.0073 \\ \hline
	\end{tabular}
\end{table}

\begin{table}[!htbp]
	\centering
	\begin{tabular}{cccc}
		\hline
		\multirow{2}{*}{Method} & \multicolumn{3}{c}{Split CIFAR100(10T)}\\ \cline{2-4}
	&ACC($ \uparrow $)	 &FM($ \downarrow $) &LA($ \uparrow $)	
\\ \hline
LWF	&0.4224$ \pm $0.0152	&0.2240$ \pm $0.0128	&0.6325$ \pm $0.0077 \\
EWC	&0.3824$ \pm $0.0440	&0.2686$ \pm $0.0473	&0.6338$ \pm $0.0092 \\
ICARL	&0.4648$ \pm $0.0061
	&0.1877$ \pm $0.0212	&0.6401$ \pm $0.0199 \\
GEM	&0.6198$ \pm $0.0135	&0.0570$ \pm $0.0104	&0.6534$ \pm $0.0166 \\
ER	&0.6322$ \pm $0.0142	&0.0582$ \pm $0.0150	&0.6687$ \pm $0.0084 \\
FTML	&0.6272$ \pm $0.0150	&0.0850$ \pm $0.0144	&0.7023$ \pm $0.0048 \\ \hline
SRKOCL-Base	&0.6320$ \pm $0.0058	&0.0652$ \pm $0.0076	&0.6727$ \pm $0.0088 \\
SRKOCL-POD	&0.6485$ \pm $0.0130	&0.0474$ \pm $0.0136	&0.6716$ \pm $0.0066 \\
SRKOCL	&0.6698$ \pm $0.0066	&0.0490$ \pm $0.0117	&0.6945$ \pm $0.0147 \\ \hline
	\end{tabular}
\end{table}

Table1 shows that our SRKOCL enjoys competitive performance on different datasets implementing different evaluation criteria. Firstly, notably, our SRKOCL outperforms other methods considered in our experiments in terms of the final average accuracy. Secondly, SRKOCL-POD have lower ACC than SRKOCL, which means the attention module would be beneficial to generate the favorable representation. That means, selecting appropriate features instead of taking all features is more reasonable. Furthermore, it is not surprised that SRKOCL-POD is better than SRKOCL-Base in terms of overall ACC and FM with the help of POD module since this mechanism can further alleviate catastrophic forgetting.

Compared to the state-of-the-art methods, we can make several observations. Firstly, traditional continual learning algorithms, such as LwF, EWC, achieve relatively poor performance on almost all datasets. Then, as a class incremental learner, ICARL model makes an improvement via a nearest-example algorithm with higher learning accuracies; however, this model has limited ability to fight against catastrophic forgetting. Furthermore, ER approach performs similarly or better than GEM algorithm in terms of ACC and FM. It is noteworthy that both of them outperform the traditional methods on different evaluation metrics since they use an episodic memory to remove forgetting. While FTML can realize the ability to learn quickly when new task arrives via meta-learning method with high LA values, this model neglects the ability of maintaining previous knowledge, which result in lower overall performance in contrast to our SRKOCL approach. Additionally, the above methods only capture a generic latent representation for each task which cannot learn to select the relevant and useful features only for the inputs of current task.

In our SRKOCL framework, an episodic memory is used to effectively avoid the catastrophic forgetting and the knowledge distillation technology is applied to make knowledge integration between the old and new knowledge which can remove the bias. Notably, our SRKOCL method outperforms all baselines considered in this paper in terms of overall ACC, confirming our discussion earlier. The results also illustrate that the SRKOCL is a promising approach for online continual learning setting.

\section{Conclusions and Future Work}

In this work, we highlight two main challenges in the field of online continual learning: catastrophic forgetting and selection of relevant knowledge. To deal with former, we use experiment replay and POD knowledge distillation strategy, where experiment replay via an episodic memory can effectively alleviate catastrophic forgetting and POD module can further make a suitable trade-off between the learning of new knowledge and the keeping of old knowledge, ultimately, further remove the forgetting due to the imbalance of the data between the current task and previous task. In addition, we use the efficient channel attention module to learn to select the appropriate knowledge according to the input from the prior knowledge instead of taking the all knowledge to further improve the performance of the model. Finally, the comprehensive experiments are performed on different benchmark datasets and our approach consistently realize promising results, and the result of ablation studies on various datasets shows the contribution of each component about the SRKOCL framework in its overall performance.

There are several directions for future work. For example, meta-learning aims to deal with learning to learn which can leverage the past knowledge learned from the previous task to accelerate adaptation to a new task, so in the future we can use meta learning technology to further improve knowledge transfer. In this paper, we have shown the effectiveness of SRKOCL in image classification area and we can consider experiments are conducted on other application domains, such as semantic segmentation, semantic detection and so on. In addition, we would also consider the online learning problem for domain incremental setting that is more complex continual learning scenario.

\bibliography{mybibfile}%摘要

\begin{thebibliography}{10}
\expandafter\ifx\csname url\endcsname\relax
  \def\url#1{\texttt{#1}}\fi
\expandafter\ifx\csname urlprefix\endcsname\relax\def\urlprefix{URL }\fi
\expandafter\ifx\csname href\endcsname\relax
  \def\href#1#2{#2} \def\path#1{#1}\fi

\bibitem{1french2006catastrophic}
R.~M. French, Catastrophic forgetting in connectionist networks, Encyclopedia
  of cognitive science (2006).

\bibitem{2sahoo2018online}
D.~SAHOO, H.~Q. PHAM, J.~LU, S.~C. HOI, Online deep learning: Learning deep
  neural networks on the fly.(2018), in: Proceedings of the Twenty-Seventh
  International Joint Conference on Artificial Intelligence IJCAI 2018, July
  13-19, Stockholm, pp. 2660--2666.

\bibitem{3pham2019bilevel}
Q.~Pham, D.~Sahoo, C.~Liu, S.~C. Hoi, Bilevel continual learning, arXiv
  e-prints (2019) 1--13.

\bibitem{4kirkpatrick2017overcoming}
J.~Kirkpatrick, R.~Pascanu, N.~Rabinowitz, J.~Veness, G.~Desjardins, A.~A.
  Rusu, K.~Milan, J.~Quan, T.~Ramalho, A.~Grabska-Barwinska, et~al., Overcoming
  catastrophic forgetting in neural networks, Proceedings of the national
  academy of sciences 114~(13) (2017) 3521--3526.

\bibitem{5zenke2017continual}
F.~Zenke, B.~Poole, S.~Ganguli, Continual learning through synaptic
  intelligence, in: International Conference on Machine Learning, PMLR, 2017,
  pp. 3987--3995.

\bibitem{6leibe2016computer}
B.~Leibe, J.~Matas, N.~Sebe, M.~Welling, Computer Vision--ECCV 2016: 14th
  European Conference, Amsterdam, The Netherlands, October 11--14, 2016,
  Proceedings, Part IV, Vol. 9908, Springer, 2016.

\bibitem{7yoon2018lifelong}
J.~YOON, E.~YANG, J.~LEE, et~al., Lifelong learning with dynamically expandable
  networks [j/ol] (2018).

\bibitem{8rusu2016progressive}
A.~A. Rusu, N.~C. Rabinowitz, G.~Desjardins, H.~Soyer, J.~Kirkpatrick,
  K.~Kavukcuoglu, R.~Pascanu, R.~Hadsell, Progressive neural networks, arXiv
  preprint arXiv:1606.04671 (2016).

\bibitem{9parisi2019continual}
G.~I. Parisi, R.~Kemker, J.~L. Part, C.~Kanan, S.~Wermter, Continual lifelong
  learning with neural networks: A review, Neural Networks 113 (2019) 54--71.

\bibitem{10douillard2020podnet}
A.~Douillard, M.~Cord, C.~Ollion, T.~Robert, E.~Valle, Podnet: Pooled outputs
  distillation for small-tasks incremental learning, in: Computer Vision--ECCV
  2020: 16th European Conference, Glasgow, UK, August 23--28, 2020,
  Proceedings, Part XX 16, Springer, 2020, pp. 86--102.

\bibitem{11rebuffi2017icarl}
S.-A. Rebuffi, A.~Kolesnikov, G.~Sperl, C.~H. Lampert, icarl: Incremental
  classifier and representation learning, in: 2017 IEEE Conference on Computer
  Vision and Pattern Recognition (CVPR), IEEE Computer Society, 2017, pp.
  5533--5542.

\bibitem{12he2020incremental}
J.~He, R.~Mao, Z.~Shao, F.~Zhu, Incremental learning in online scenario, in:
  Proceedings of the IEEE/CVF Conference on Computer Vision and Pattern
  Recognition, 2020, pp. 13926--13935.

\bibitem{13robins1993catastrophic}
A.~Robins, Catastrophic forgetting in neural networks: the role of rehearsal
  mechanisms, in: Proceedings 1993 The First New Zealand International
  Two-Stream Conference on Artificial Neural Networks and Expert Systems, IEEE,
  1993, pp. 65--68.

\bibitem{14lopez2017gradient}
D.~Lopez-Paz, M.~Ranzato, Gradient episodic memory for continual learning,
  Advances in neural information processing systems 30 (2017) 6467--6476.

\bibitem{15chaudhry2018riemannian}
A.~Chaudhry, P.~K. Dokania, T.~Ajanthan, P.~H. Torr, Riemannian walk for
  incremental learning: Understanding forgetting and intransigence, in:
  European Conference on Computer Vision, Springer, 2018, pp. 556--572.

\bibitem{16he2016deep}
K.~He, X.~Zhang, S.~Ren, J.~Sun, Deep residual learning for image recognition,
  in: Proceedings of the IEEE conference on computer vision and pattern
  recognition, 2016, pp. 770--778.

\bibitem{17sokar2021self}
G.~Sokar, D.~C. Mocanu, M.~Pechenizkiy, Self-attention meta-learner for
  continual learning, in: Proceedings of the 20th International Conference on
  Autonomous Agents and MultiAgent Systems, 2021, pp. 1658--1660.

\bibitem{18wang2020eca}
Q.~Wang, B.~Wu, P.~Zhu, P.~Li, W.~Zuo, Q.~Hu, Eca-net: Efficient channel
  attention for deep convolutional neural networks, in: 2020 IEEE/CVF
  Conference on Computer Vision and Pattern Recognition (CVPR), IEEE Computer
  Society, 2020, pp. 11531--11539.

\bibitem{19ioannou2017deep}
Y.~Ioannou, D.~Robertson, R.~Cipolla, A.~Criminisi, Deep roots: Improving cnn
  efficiency with hierarchical filter groups, in: 2017 IEEE Conference on
  Computer Vision and Pattern Recognition (CVPR), IEEE, 2017, pp. 5977--5986.

\bibitem{20krizhevsky2009learning}
A.~Krizhevsky, G.~Hinton, et~al., Learning multiple layers of features from
  tiny images (2009).

\bibitem{21rajasegaran2020itaml}
J.~Rajasegaran, S.~Khan, M.~Hayat, F.~S. Khan, M.~Shah, itaml: An incremental
  task-agnostic meta-learning approach, in: 2020 IEEE/CVF Conference on
  Computer Vision and Pattern Recognition (CVPR), IEEE, 2020, pp. 13585--13594.

\bibitem{22chaudhry2019continual}
A.~Chaudhry, M.~Rohrbach, M.~Elhoseiny, T.~Ajanthan, P.~K. Dokania, P.~H. Torr,
  M.~Ranzato, Continual learning with tiny episodic memories (2019).

\bibitem{23finn2019online}
C.~Finn, A.~Rajeswaran, S.~Kakade, S.~Levine, Online meta-learning, in:
  International Conference on Machine Learning, PMLR, 2019, pp. 1920--1930.

\bibitem{24riemer2018learning}
M.~Riemer, I.~Cases, R.~Ajemian, M.~Liu, I.~Rish, Y.~Tu, G.~Tesauro, Learning
  to learn without forgetting by maximizing transfer and minimizing
  interference, arXiv preprint arXiv:1810.11910 (2018).

\end{thebibliography}

\end{document}